\documentclass[lettersize,journal]{IEEEtran}
\usepackage{amsmath,amsfonts}
\usepackage{algorithmic}
\usepackage{algorithm}
\usepackage{array}
\usepackage[caption=false,font=normalsize,labelfont=sf,textfont=sf]{subfig}
\usepackage{textcomp}
\usepackage{stfloats}
\usepackage{url}
\usepackage{verbatim}
\usepackage{graphicx}
\usepackage{cite}
\usepackage{tabularx}
\usepackage{booktabs}
\begin{document}




\title{The impact of objective interactions on the performance of massive objective optimization algorithms}

\author{Shakiba Shahbandegan, Jose Guadalupe Hernandez, and Emily Dolson
}

\markboth{Journal of \LaTeX\ Class Files,~Vol.~14, No.~8, August~2021}%
{Shell \MakeLowercase{\textit{et al.}}: A Sample Article Using IEEEtran.cls for IEEE Journals}


\maketitle

\begin{abstract}
Many-objective optimization has been a field of interest over the past two decades and several evolutionary optimization algorithms have been introduced to tackle these problems; yet two fundamental questions remain underexplored: (i) What happens when the number of objectives grows beyond the typical many-objective regime of about fifteen and becomes massive? (ii) How do problem characteristics, such as the nature of interactions between objectives, influence algorithmic performance? To answer these questions we employ a diagnostic benchmark suite that allows control over problem characteristics and can be scaled to extremely high objective counts. Using this framework we evaluate several state-of-the-art evolutionary algorithms including NSGA-II, NSGA-III, MOEA/D and lexicase selection across a range of dimensionalities and diagnostic problem landscapes. Our experiments reveal that problem characteristics significantly affect algorithm performance. In particular, the nature of interactions between objectives appears important. These results highlight the importance of understanding these properties before selecting an algorithm for a specific problem. We also show that lexicase selection, an algorithm originally designed for genetic programming, compares favorably with state-of-the-art many-objective optimization algorithms while avoiding the dependence on predefined reference directions. 

\end{abstract}


\section{Introduction}
\IEEEPARstart{R}{eal}-world optimization problems involve the simultaneous improvement of multiple objectives and arise in multiple domains such as bioinformatics \cite{ref4}, software engineering \cite{ref1}, multi-agent systems \cite{ref3}, and automated machine learning \cite{tpot}. 
When the number of objectives is limited to two or three, the problem is typically referred to as a multi-objective optimization problem (MOP), whereas problems with more than three objectives are classified as many-objective optimization problems (MaOP). 
Navigating excessively high-dimensional objective spaces is particularly challenging due to the need to balance numerous trade-offs and the interactions among them.
These challenges reduce the effectiveness of conventional optimization methods, as the computational cost of evaluating and comparing solutions grows exponentially, and the structure of higher-dimensional Pareto fronts become increasingly complex \cite{NSGA3}.
Furthermore, many performance indicators that assess solution quality based on convergence and diversity (e.g. hypervolume measures) become too costly to calculate. 

Over the past two decades, evolutionary many-objective optimization methodologies have been developed to address many-objective optimization problems. 
Although these algorithms have shown promising performance, several aspects remain unexplored in the literature. 
One major limitation in prior research is that many-objective optimization is typically limited to up to 15 objectives\cite{ref5,ref6}; 
relatively little work has been done on problems involving massive objectives due to several reasons. Studying massive objective problems comes with very high computational cost as more objectives result in more evaluation functions, multiplying the computational burden of selection, ranking, and visualization of candidate solutions. In addition, most benchmark suites such as ZDT \cite{ZDT}, DTLZ \cite{DTLZ}, and WFG \cite{WFG} are not designed for settings with extremely high numbers of objectives. One might argue that such high-dimensional problems are rare in practice and not worth the costs;
yet many real-world problems could benefit from a massive-objective formulation if robust optimization techniques were available and their capabilities and limitations were well understood.
For instance, in fair machine learning, one approach to mitigating algorithmic bias is to build classifiers that optimize trade-offs across multiple demographic subgroups. 
Depending on the data set and the categorization method, the number of subgroups---and therefore the number of objectives---can easily exceed 15 \cite{margint}. 

The interactions between the objectives are another key aspect that has been largely overlooked in the literature.
Most studies assume that objectives are inherently contradictory, meaning that improving one objective necessarily leads to the deterioration of others. 
However, the relationship between the objectives can vary, and be categorized along the following continuum \cite{gecco24}:

\noindent \textbf {Synergistic}: An improvement in one objective simultaneously enhances one or more objectives. 
For example, in aerodynamic vehicle design, optimizing the body shape to reduce air resistance (drag) often results in improved fuel efficiency. 

\noindent \textbf {Orthogonal}: The performance on one objective is independent of all others, meaning that improvements in one objective do not impact the rest. 
For instance, in a vehicle purchase, paint-color preference is orthogonal to fuel economy---maximizing satisfaction with color does not affect fuel efficiency, and vice versa.

\noindent \textbf {Contradictory}: Enhancing one objective results in the deterioration of one or more objectives. 
For example, ensuring higher quality in manufacturing a product often leads to an increase in costs. 

Despite their practical significance ---since such interactions frequently arise in real-world applications---, the relationships between objectives have not been thoroughly examined in the context of multi and many-objective optimization problems. 
Understanding how these relationships influence performance is important for developing more robust and scalable optimization algorithms, especially when a large number of objectives are involved. 

This study aims to address these gaps by systematically analyzing the impact of objective interactions, other problem characteristics, and scaling effects in multi- and many-objective optimization problems. 
For this purpose, we employ the DOSSIER benchmark suite introduced in \cite{diags}, a handcrafted set of test problems designed to measure the search capacities of evolutionary algorithms that encode different patterns of objective interactions and gives us flexibility to scale the number of objectives for each test problem. 
We select a variety of multi- and many-objective optimization algorithms and evaluate their performance on these landscapes. 
The first of these algorithms are the current state-of-the-art many-objective optimization algorithms: NSGA-III \cite{NSGA3} and MOEA/D \cite{MOEAD}. 
We also examined NSGA-II \cite{NSGA2}, which was originally developed for multi-objective optimization (less than 4 objectives) but has shown to be competitive for larger number of objectives \cite{NSGA2>NSGA3, NSGA2>NSGA3_2, NSGA2>NSGA3_3}. 
We also included lexicase selection \cite{lex2012} and its variants \cite{elex} in our experiments. 
Lexicase is a parent selection technique originally developed for genetic programming; therefore, it has not been fully investigated for use in multi-objective optimization algorithms. 
However, recent studies have suggested that lexicase could be effective in such problems \cite{gecco24, bill-poster}. 

The remainder of this paper is structured as follows: Section II details the methodology, including the benchmark suite and selected algorithms for the experiments. Section III reviews previous work on many-objective optimization and algorithms developed to address these problems. Section IV describes the experimental setup of this work. Section V presents the results, highlighting comparative algorithm performance on benchmark problems. Finally, Section VI discusses and summarizes key findings and outlines potential directions for future research.

\section{Methods}

\subsection{Benchmark}

A variety of benchmark problems are available for many-objective optimization that provide insights into algorithm performance. Among the most commonly used are the ZDT \cite{ZDT}, DTLZ \cite{DTLZ}, and WFG \cite{WFG} test problems. 
These benchmarks provide well-defined analytical objective functions that emphasize the geometry of the Pareto front and are primarily designed to evaluate scalability, convergence, and diversity preservation of optimization algorithms. However, they become difficult to interpret in dimensions higher than 3, and they offer limited insight into the underlying interactions between objectives and decision variables. 

To address these limitations, we adopt the DOSSIER benchmark suite (Diagnostic Overview of Selection Schemes in Evolutionary Runs) \cite{diags}.
The DOSSIER diagnostic suite contains a set of test problems designed to assess an algorithm's strengths and weaknesses by measuring its performance on search spaces with varying characteristics.  The test problems all follow the same pattern. A vector of \textit{D} floating point values (the genotype) serves as input. By default, each genotype contains 100 values in [0, 100] range, but these parameters can be modified to change the difficulty of the problem. The genotype is translated into a numerical vector of ``objective scores'' with the same dimensionality. Each problem in the DOSSIER suite specifies a different rule for how this translation happens. Every position in the resulting objective scores vector represents the score on one objective in the overall multi-objective optimization problem. As a result, the number of variables and the number of objectives are the same in each problem. For each diagnostic, the set of all optimal non-dominated solutions is known and is referred to as the reference Pareto front; this reference front will be important for quantpeach fallsifying performance. More details about each diagnostic are discussed below, and summarized in Table 1.

Importantly for our purposes, the objectives in each of these problems interact with each other in different ways. We quantified these interactions by 1) generating a random population of genotype vectors, 2) applying random single-gene mutations to those vectors, and then 3) quantifying the effect that each mutation had on each objective score. If increasing the gene value increased the value of an objective score, that mutation was classified as synergistic with that objective. If the mutation decreased the value of the objective score, it was classified as contradictory with that objective. If it had no effect, it was classified as orthogonal with that objective. Results of this analysis are shown in the pie charts in Table 1. Note that these results include ``self interactions", \textit{i. e.} the effect that mutating a gene has on its own corresponding objective.

The diagnostics in the DOSSIER suite are as follows:

\textbf{Exploitation Rate}: In this diagnostic, a genotype is mapped directly to its vector of objective scores without any modifications, implying no interactions between the objectives (i.e. the objectives are orthogonal). 
The goal is to maximize each objective independently. 
The reference Pareto front for this test problem consists of a single solution in which all objectives are simultaneously maximized, representing the global optimum. 
The following equation shows how a genotype \textit{x} is translated to its objective score:

\begin{equation}
O_i = x_i, \quad \forall i \in \{1, 2, \dots, D\}
\end{equation}

\noindent where $x_i$ is the value of the i-th position in the genotype and $O_i$ is its corresponding objective score. Based on this equation, every time a variable $x_i$ is mutated in the genotype, it affects only its corresponding objective in the phenotype (i.e. it is synergistic with itself; see blue portion of pie chart in Table 1) and affects no other objective (orthogonal to all other objectives, color gray in the pie chart in Table 1). 

\textbf{Multi-path Exploration}: As the name suggests, this diagnostic evaluates an algorithm's ability to explore multiple paths through the search space simultaneously. The translation function identifies the highest value in the vector and retains it, along with all subsequent values that are smaller than their immediate predecessor. Every other value is set to zero. Equation (2) represents the mathematical formulation for this test problem.

\begin{equation}
O_i =
\begin{cases}
x_i, & \text{if } i = m \\
x_i, & \text{if } i > m \text{ and } x_i < x_{i-1} \text{ and } O_{i-1} \neq 0 \\
0,   & \text{otherwise}
\end{cases}
\end{equation}

\noindent 
where $m$ is the position of the highest value in the genotype.
The Multi-path Exploration diagnostic creates a mix of synergy and contradiction between the objectives. 
Synergy occurs because improving one objective can lead to improvements in subsequent objectives with smaller values. 
However, improving one objective can also cause other values to be set to zero, creating contradiction. Any mutation that does not affect the current highest value in the genotype will be orthogonal, which explains the larger gray area in the pie chart in Table 1.

\textbf{Contradictory Objectives}: This diagnostic evaluates an algorithm’s ability to identify multiple global optima in a scenario where objectives are in some degree of conflict. Improving one objective too much worsens all others. 
The translation function retains only the gene with the highest value in the genotype, setting all other values to zero, as shown in equation (3). 
Consequently, any objective score where a single position is maximized while all others are zero qualifies as a solution on the reference Pareto front. There are $D$ (the problem's dimensionality) such solutions.

\begin{equation}
O_i =
\begin{cases}
x_i, & \text{if } i = m \\
0,   & \text{otherwise}
\end{cases}
\end{equation}

\noindent 
where $m$ is the position of the highest value in the genotype.
In this diagnostic, most mutations are orthogonal, unless a mutation causes a new gene to surpass the current maximum in the genotype. When this occurs, the affected variable acts synergistically with its corresponding objective. Meanwhile, it acts contradictorily with the previously dominant gene, as its associated objective score drops to zero. Thus, only mutations that shift the position of the maximum gene induce meaningful objective interactions; all others are neutral. 

\textbf{Diversity}: Similar to the contradictory objectives diagnostic, the diversity diagnostic creates a scenario where only one objective score can be maximized at a time. 
However, whereas under the contradictory objectives diagnostic most mutations have a neutral effect, the diversity diagnostic creates a fitness gradient that results in
contradiction between the mutated objective and the current highest objective. 
In the Diversity diagnostic, the objective score for the gene with the maximum value is that value itself. The objective score for every other gene is computed by subtracting that gene from the maximum gene value, and then halving the result. 
This transformation creates selective pressure to maximize a single trait while minimizing the others. 
As a result, the reference Pareto front consists of solutions where one objective is maximized and all remaining objectives have values equal to half of that maximum. Equation (4) formulates this test problem.


\begin{equation}
\begin{cases}
x_i, & \text{if } i=m \\
\frac{1}{2} \left( x_m - x_i \right), & \text{otherwise}
\end{cases}
\end{equation}

\noindent where $m$ is the position of the highest value in the genotype. Looking at the pie chart for this diagnostic in Table 1, we can see that it produces substantially more contradictory objectives than the previous diagnostics and also a high percentage of synergistic interactions. 

\textbf{Antagonistic Contradictory Objectives Diagnostic}: This diagnostic was proposed in \cite{gecco24} and extends the Contradictory Objectives diagnostic by intensifying conflicts between the objectives. 
The objective score is derived from its corresponding genotype value minus the sum of all other genes. 
Consequently, an improvement in one objective will maximally negatively affect all other objectives. Equation (5) formulates the fitness function for this diagnostic:

\begin{equation}
    O_i = x_i - \sum\limits_{\substack{j=0 \\ j \neq i}}^{D} x_j
\end{equation}

The reference Pareto front for this diagnostic consists of all genotypes where a single position is nonzero and all other positions are zero. 
Any mutation in any of the values in the genotype will have an affect on all the values in the phenotype. As a result, there is no orthogonality in this diagnostic. The mutated value will be synergistic to itself, while being contradictory to every other value.

\begin{table*}
    \centering
    \caption{
        Summary of the diagnostic fitness functions. The pie charts show objective interactions in three colors; blue, yellow and gray represent synergistic, contradictory and orthogonal interactions respectively. 
    }
    \label{tbl:diagnostics}
    \begin{tabular}{|m{3cm}|m{4cm}|m{4cm}|m{4cm}|}
\hline

\textbf{Diagnostic} & \textbf{Example genotype phenotype mapping} & \textbf{Reference PF in phenotype space} & \textbf{Objective effects} \\
\hline

Exploitation Rate &
\begin{tabular}[c]{@{}l@{}}genotype: {[}0, 5, 10, 4, 1{]}\\ phenotype: {[}0, 5, 10, 4, 1{]}\end{tabular} &
{[}10, 10, 10, 10, 10{]} &
\includegraphics[width=4.2cm]{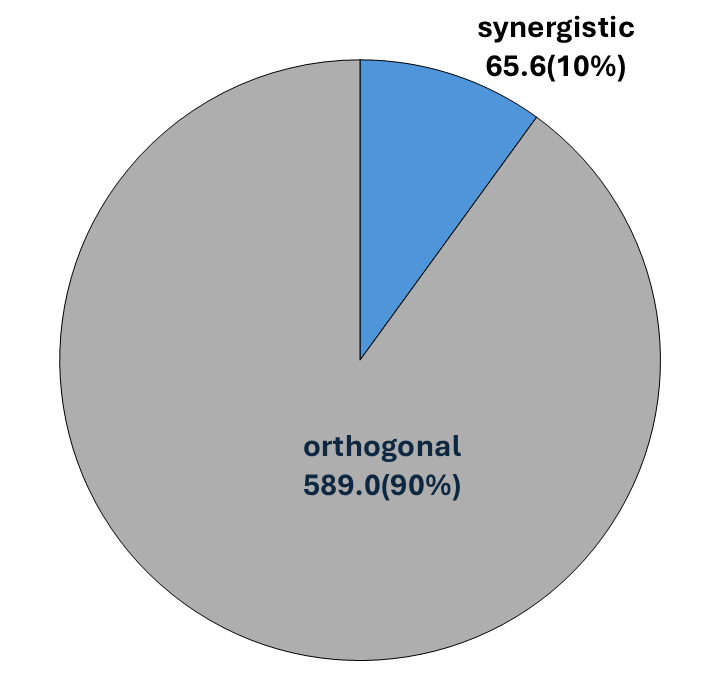} \\
\hline

Exploration Rate &
\begin{tabular}[c]{@{}l@{}}genotype: {[}0, 5, 10, 4, 1{]}\\ phenotype: {[}0, 0, 10, 4, 1{]}\end{tabular} &
{[}10, 10, 10, 10, 10{]} &
\includegraphics[width=4.2cm]{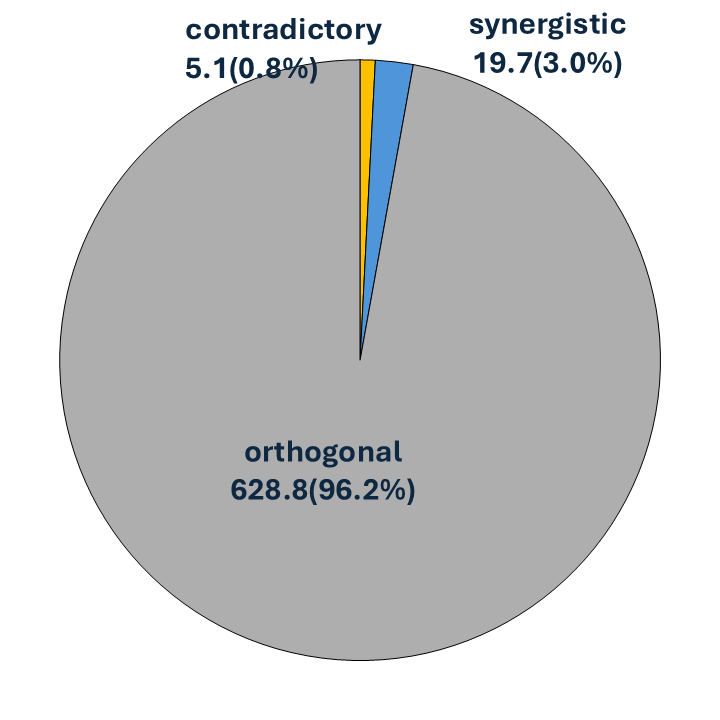} \\
\hline

Contradictory Objectives &
\begin{tabular}[c]{@{}l@{}}
genotype: {[}0, 5, 10, 4, 1{]} \\
phenotype: {[}0, 0, 10, 0, 0{]}
\end{tabular} &
\begin{tabular}[c]{@{}l@{}}
{[}10, 0, 0, 0, 0{]} \\
{[}0, 10, 0, 0, 0{]} \\
{[}0, 0, 10, 0, 0{]} \\
{[}0, 0, 0, 10, 0{]} \\
{[}0, 0, 0, 0, 10{]}
\end{tabular} &
\includegraphics[width=4.2cm]{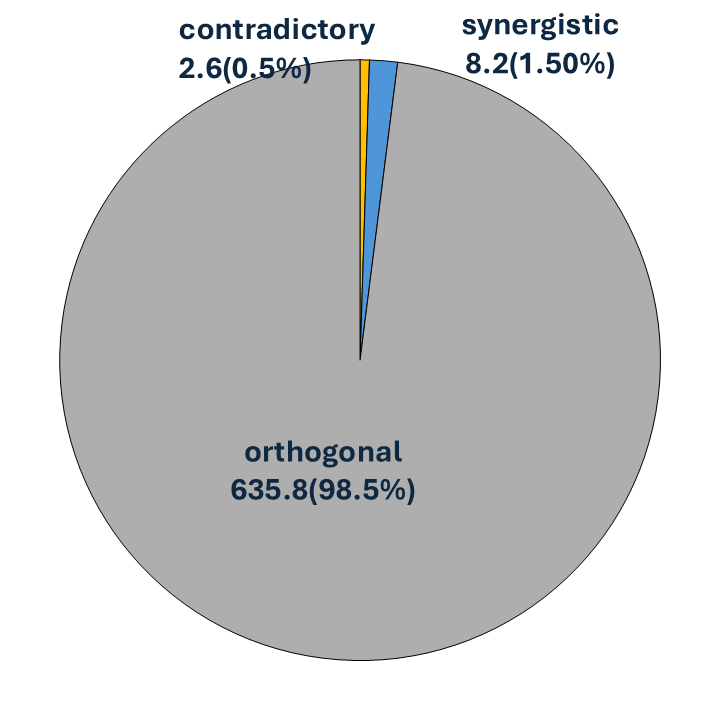} \\
\hline

Diversity &
\begin{tabular}[c]{@{}l@{}}genotype: {[}0, 5, 10, 4, 1{]}\\ phenotype: {[}5, 2.5, 10, 3, 4.5{]}\end{tabular} &
\begin{tabular}[c]{@{}l@{}}
{[}10, 5, 5, 5, 5{]} \\
{[}5, 10, 5, 5, 5{]} \\
{[}5, 5, 10, 5, 5{]} \\
{[}5, 5, 5, 10, 5{]} \\
{[}5, 5, 5, 5, 10{]}
\end{tabular} &
\includegraphics[width=4.2cm]{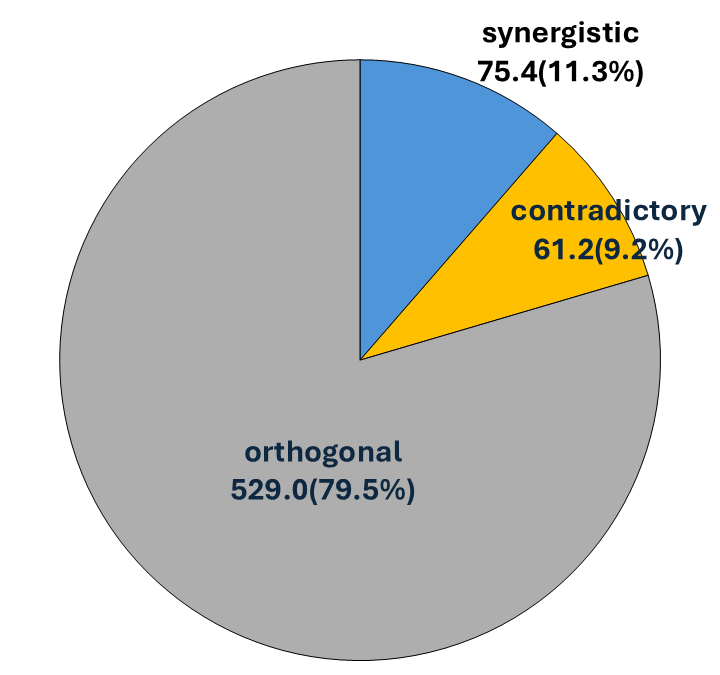} \\
\hline

Antagonistic Contradictory Objectives &
\begin{tabular}[c]{@{}l@{}}genotype: {[}0, 5, 10, 4, 1{]}\\ phenotype: {[}-20, -10, 0, -12, -18{]}\end{tabular} &
\begin{tabular}[c]{@{}l@{}}
{[}*, 0, 0, 0, 0{]} \\
{[}0, *, 0, 0, 0{]} \\
{[}0, 0, *, 0, 0{]} \\
{[}0, 0, 0, *, 0{]} \\
{[}0, 0, 0, 0, *{]}
\end{tabular} &
\includegraphics[width=4.2cm]{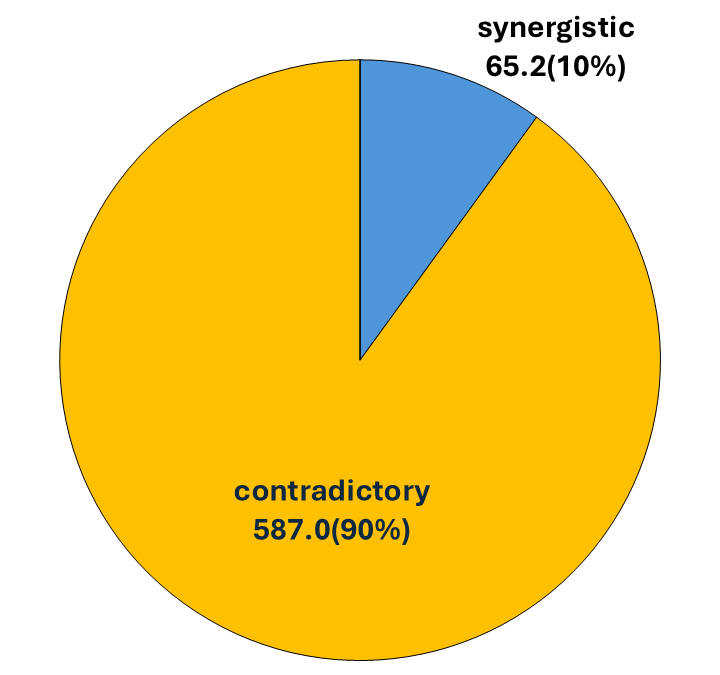} \\
\hline

\end{tabular}

\end{table*}


\subsection{Algorithms}

\textbf{NSGA-II:} Non-dominated Sorting Genetic Algorithm II (NSGA-II) \cite{NSGA2} is a widely used evolutionary algorithm, originally designed for solving multi-, rather than many- objective optimization problems. NSGA-II identifies a set of Pareto optimal solutions by ranking candidate solutions based on Pareto dominance; a solution is said to dominate another solution if it is strictly better than that solution on at least one objective and no worse on the others. The algorithm sorts the individuals in the population and assigns them to different fronts. The first front consists of solutions not dominated by any other solution, and subsequent fronts contain solutions dominated only by those in the previous fronts. To maintain diversity, NSGA-II ranks solutions within a front using a niche-preservation operator called crowding distance, which calculates the density of solutions surrounding each candidate. Individuals in less crowded regions of the solution space are favored to ensure a well-spread Pareto front. This ranking is then used for parent selection, which is achieved via binary tournament selection. Resulting offspring are added to the population. Finally, at the end of each generation, Pareto ranking is used to cull the population down to the desired size. 

NSGA-II has been proven successful in problems with 2-3 objectives. However, there has been limited research on its performance on many-objective optimization problems. In several studies NSGA-II was found to outperform other state-of-the-art many-objective optimization algorithms on certain problems \cite{NSGA2>NSGA3, NSGA2>NSGA3_2, NSGA2>NSGA3_3}. Therefore, we include NSGA-II in our experimental comparisons to provide a more comprehensive performance assessment.

\textbf{NSGA-III}: Non-dominated Sorting Genetic Algorithm III (NSGA-III) \cite{NSGA3} is similar to NSGA-II in many ways. 
Based on non-dominated sorting, it divides the population into different fronts and assigns them ranks. 
Parent selection in NSGA-III is typically random, with all optimization pressure coming from the survival selection at the end of each generation. 
The key distinction lies in its approach to maintaining diversity. 
While NSGA-II uses the crowding distance metric, NSGA-III promotes diversity by selecting solutions based on their proximity to a set of predefined reference directions. 
These reference directions guide the selection process to ensure that solutions are well distributed. To create the reference directions, each objective is first divided into \textit{p} partitions. 
The total number of reference points in a problem with $M$ objectives (\textit{H}) is then given by:

\begin{equation}
 H = \binom{M + p - 1}{p}
\end{equation}

It is important in NSGA-III that the population size is at least as big as the number of reference points. That way, at least one population member is associated with each reference direction. 
In the original paper it is recommended that the population size be equal to the smallest multiple of four bigger than the number of reference points. 

When the number of objectives exceeds eight, the number of reference points become extremely large. 
To avoid that, the designers of NSGA-III recommend a multi-layer approach \cite{NSGA3} with relatively smaller values of \textit{p} in each layer. The first layer forms a coarse set of well-spread reference points on the unit simplex, and subsequent layers add additional points that fill the gaps between the outer layer and the center. By stacking these layers, NSGA-III achieves good coverage of the objective space without requiring a single, extremely fine discretization, thereby keeping the total number of reference points computationally manageable.
In our experiments, we use this approach for problems with up to 15 objectives. For problems with more than 15 objectives we used a linear five layer approach based on the guidlines in \cite{NSGA3-exeter}.

\textbf{MOEA/D}: Multi-objective Evolutionary Algorithm Based on Decomposition (MOEA/D) \cite{MOEAD}, as the name suggests, decomposes a many objective problem into multiple single objective subproblems, each corresponding to a particular weight vector. 
MOEA/D associates each subproblem with a different weight vector. 
For each weight vector, \textit{T} weight vectors are defined as neighbors based on Euclidean distance. 
In the initialization step, one individual is randomly assigned to a weight vector. Then two individuals among the neighbors of each subproblem are selected to exchange genetic material. 
An offspring is created whose fitness is evaluated using one of the decomposition methods, and is assigned to a neighboring weight vector accordingly. 
Decomposition methods used for MOEA/D include weighted sum, Tchebycheff, and penalty-based boundary intersection (PBI) \cite{MOEAD}. 
In our experiments, the PBI decomposition method is used. PBI is a scalarization method that evaluates how well a solution aligns with a specific weight vector, balancing convergence toward the Pareto front with diversity across the front. 
For solution $x$, PBI is calculated via the following equation:

\begin{equation}
\text{PBI(x)} = d_1 + \theta \cdot d_2
\end{equation}

\noindent where for a given weight vector, $d_1$ denotes the $x$'s distance along the direction of the weight vector and $d_2$ is the perpendicular distance of $x$ to the weight vector. $\theta$ is a parameter that controls the balance between convergence and diversity. 

\textbf{Lexicase}:
Lexicase is a parent selection algorithm originally developed for genetic programming \cite{lex2012}. 
In genetic programming, solutions are commonly evaluated by running them on various test cases and comparing their output to the expected output. 
To select a parent, lexicase steps through the test cases in a randomly shuffled order. For each test case, the individual(s) with the best fitness are kept in consideration for selection, while the others are eliminated. 
This process continues until only one individual is left in the selection pool; this individual is selected to reproduce. 
If no test case is left and more than one individual is in the selection pool, one individual is selected randomly. 
While it was not originally designed for multi-objective optimization, lexicase selection can be used for this purpose by treating each objective as a test case \cite{elex}. 
In the context of the diagnostic test problems used in this paper, an individual is a genotype and each gene/objective is a test case.

If the genes in an individual contain floating point values, it is rare for multiple individuals to tie for having the best performance on a given test case. 
As a result, selection would be based exclusively on the first objective randomly chosen, which is generally expected to produce undesirable behavior. 
To solve this problem, $\varepsilon$-lexicase is introduced in \cite{elex}. 
$\varepsilon$-lexicase follows the same procedure as standard lexicase, with one key difference: instead of retaining only the individual with the best fitness on each objective, it retains all individuals whose fitness values are within an $\varepsilon$ threshold of the best. 
The value of $\varepsilon$ is an algorithmic parameter and can be a constant floating point value depending on the problem. 
However, in \cite{elex} it is suggested that the value of $\varepsilon$ be automatically selected based on the following equation:

\begin{equation}
    \varepsilon(j) = median(|O_j(n_i) - median(O_j(n_i))|) \forall n_i \in N
\label{eq:medianepsilon}
\end{equation}
where $O_j(n_i)$ is the value of objective $O_j$ for individual $n_i$ in population $N$. There are multiple subtle variations of the $\varepsilon$-lexicase algorithm. We consider the following two:

\noindent \textbf{Semi-dynamic:} Once every generation, before the selection process begins, $\varepsilon$ is calculated for each test case across all the members in the population.

\noindent \textbf{Dynamic:} During the selection process, $\varepsilon$ is calculated relative to the current selection pool after every filtering step. 

Note that lexicase selection is just a parent selection algorithm, whereas the other algorithms discussed here also have survival selection components. Thus, our experiments here may put lexicase at a slight disadvantage, as it could potentially be combined with a survival selection component for additional improvement. 

\section{Literature Review}

\subsection{NSGA-II}

NSGA-II was originally developed for multi-objective optimization (less than 4 objectives) and several studies have reported its poor performance on larger numbers of objectives \cite{10.5555/1762545.1762607, 10266760}. These studies suggest that the primary challenge arises from the increased proportion of non-dominated solutions as the number of objectives grows, resulting in a large fraction—or even the entirety—of the population being assigned to the first Pareto front. Consequently, the responsibility for distinguishing between solutions shifts almost entirely to the crowding distance operator, which was originally intended as a secondary diversity-preserving mechanism. Previous work has introduced extensions and modifications to NSGA-II to address these limitations and better scale it with increasing numbers of objectives.

Elarbi \textit{et. al} \cite{7866900} proposed the reference-point-dominance-based (RPD) NSGA-II, where solutions are sorted and ranked into different fronts based on their distances to pre-defined reference directions, with the dominated solutions being assigned a distance of 0. Instead of using the crowding distance operator, the last front is truncated by preferring solutions that have the minimum distance in their corresponding reference directions. Experiments on DTLZ and WFG test problems with 5 to 20 objectives show competitive performance of RPD NSGA-II compared to state-of-the-art many-objective optimization algorithms such as NSGA-III and MOEA/D. 

Köppen and Yoshida \cite{10.1007/978-3-540-70928-2_55} introduced different distance assignment approaches to replace the crowding distance operator. These distance metrics quantify the degree by which a solution $A$ is dominated by solution $B$ based on the number and magnitude of smaller (or larger) objectives. They compared the original NSGA-II using a crowding distance operator with each distance assignment on DTLZ problems with 2 to 15 objectives and concluded that the $\varepsilon$-dominance distance is a promising strategy to make NSGA-II more suitable for many-objective optimization problems. 

Pan \textit{et. al} \cite{9229403} showed that a simple modification to the fitness function can enhance NSGA-II's performance on many-objective optimization problems significantly. Rather than changing the algorithm’s structure, their approach introduces a variable $\alpha$ into the fitness function which reshapes the Pareto front, encouraging dominance-resistant solutions—-those typically found on the periphery of the objective space-—to shift toward the center of the feasible region. As a result, these solutions become more susceptible to being dominated, thereby improving selection pressure and facilitating better convergence in many-objective scenarios. Experimental results on 10 objective test problems from DTLZ, WFG and HTNY show that this modification improves the convergence ability of NSGA-II. 

While these studies suggest that NSGA-II struggles with many-objective optimization problems, other research has demonstrated scenarios where NSGA-II not only performs competitively but even outperforms algorithms specifically designed for handling a larger number of objectives.

Li \textit{et. al} \cite{NSGA2>NSGA3_2} performed a comparative study between many-objective optimization problems on DTLZ, WFG, TSP and Pareto Box test problems with 5 and 10 objectives. Interestingly, results showed that NSGA-II does not always perform badly on these problems. In fact, in 5-objective DTLZ7 and WFG8 test problems, NSGA-II outperformed state-of-the-art many-objective optimization problems such as MOEA/D and HypE. 

Ishibuchi \textit{et. al} \cite{NSGA2>NSGA3} compared the performance of NSGA-II and NSGA-III on DTLZ, Distance \cite{distance_test_problems}, and Knapsack test problems with 3 to 10 objectives and concluded that NSGA-III does not always outperform NSGA-II in many-objective optimization problems; instead, the results depend on the number of objectives and the characteristics of the test problems. Notably, NSGA-II outperformed NSGA-III on the Knapsack problem, while NSGA-III demonstrated superior performance on DTLZ.

In another study, Ishibuchi \textit{et. al} \cite{NSGA2>NSGA3_3} generated 18 types of test problems by combining different properties of Pareto fronts and feasible regions. They then concluded that, NSGA-II outperformed the other examined EMO algorithms on a half of the 18 types of test problems with 8 and 10 objectives. 

These studies have challenged the assumption that NSGA-II does not perform well in many-objective scenarios, and show that its performance seems to be highly problem dependent.

\subsection{NSGA-III}

In the original NSGA-III study \cite{NSGA3}, the algorithm is extensively investigated on a large number of problems with 3 to 15 objectives. These problems have various types of Pareto fronts such as concave, convex, disjoint, differently scaled, multimodal (i.e containing multiple local fronts), and with biased density of points across the front. In contrast, while multiple variants of MOEA/D were evaluated, no single version demonstrated the ability to solve all problems efficiently.

Saeda \textit{et. al} \cite{UNSGA3} proposed a unified version of NSGA-III (UNSGA-III) capable of solving not only many-, but also multi- and mono-objective optimization problems. Based on the input, this unified approach automatically chooses an efficient population-based algorithm tailored to each problem class. To select parent solutions, UNSGA-III uses a niching-based tournament selection operator that operates by randomly drawing two solutions from the population. If these solutions are associated with different reference directions, a winner is selected randomly. If the solutions share the same reference direction, the one with a better Pareto rank is chosen. If the solutions belong to the same rank and niche, the one closest to the reference direction is selected. The performance of UNSGA-III was evaluated on unconstrained and constrained benchmark problems with a range of 1 to 10 objectives, including bi-objective test cases, the ZDT and DTLZ suites, and two real-world engineering problems. The study concluded that UNSGA-III performs comparably to, and in some cases outperforms, its respective counterpart: an elite-preserving Genetic Algorithm for single-objective problems, NSGA-II for bi-objective problems, and NSGA-III for three- and many-objective problems. 

Vesikar \textit{et. al} \cite{RNSGA3} proposed the Reference-Point-Based NSGA-III (RNSGA-III), another variant of NSGA-III that focuses on finding a specific preferred region of the Pareto front. This algorithm introduces a novel reference point generation method based on user-defined "aspiration points", but otherwise uses the same genetic operators, selection, and survival mechanisms as NSGA-III. The performance of RNSGA-III was tested on DTLZ and WFG problems with up to 5 objectives as well as an engineering problem. Results on these problems demonstrated that RNSGA-III effectively identifies Pareto-optimal solutions within regions of interest. In addition, it can validate if there exists any Pareto-optimal point in regions where a former algorithm failed to find any solution.

To improve the convergence capability of NSGA-III, Cui \textit{et. al} \cite{NSGA3-SE} introduced new selection and elimination operators (NSGA-III-SE). The selection process chooses the individual with the minimum niche count and shortest PBI distance, while the elimination process removes the individual with the maximum niche count and longest PBI distance. This algorithm was evaluated on DTLZ and WFG problems with 3 to 15 objectives, and compared to five other evolutionary algorithms. Experimental results showed that overall, NSGA-III-SE achieved solutions exhibiting both strong convergence and high diversity, and outperformed all other algorithms considered, with the exception of one MOEA/D variant.

Liu \textit{et. al} \cite{NSGA3-GKM} proposed an enhancement to NSGA-III by incorporating a genetic K-means clustering algorithm. This approach clusters the initial reference points and replaces them with the corresponding cluster centers. By partitioning the objective space into distinct subspaces and introducing a PBI-based aggregation function, the modified algorithm aims to improve both convergence and diversity of the solutions. The proposed method was evaluated on the DTLZ and UF benchmark suites with up to 10 objectives, and had the best average performance based on the IGD and HV indicators compared to various NSGA-II variants.

One downside of NSGA-III is that generating reference directions becomes impractical when the number of objectives becomes large (e.g., for eight objectives, the number of reference points can get as large as 5040 even with only one partition). To solve this problem, in the original NSGA-III paper, the authors recommended a multilayer reference point generation technique to limit the number of reference points and hence the population size \cite{NSGA3}. However, a major assumption in this method is that the reference directions have a uniform distribution, which might not be the case in problems with a massive number of objectives. To address this limitation, Li \textit{et. al} proposed using non-linear shrinking factors to generate the reference directions \cite{NSGA3-exeter}. They also derived mathematical formulations to determine optimal non-linear factors. The authors applied this method of generating reference directions to three evolutionary algorithms: NSGA-III, RNSGA-II and MOEA/D and. They compared the performance of these algorithms on DTLZ problems with up to 100 objectives and concluded that RNSGA-III is the best performing candidate on all benchmark problems compared to NSGA-III and MOEA/D. However, their experiments did not consider alternative reference point generation methods for the same algorithms, nor did they include evolutionary algorithms that operate without reference point mechanisms.

\subsection{MOEA/D}
Another famous algorithm in many-objective optimization is the decomposition-based algorithm MOEA/D, described in Section II. Multiple studies in the literature have investigated enhancements to MOEA/D. Qi \textit{et. al} \cite{MOEAD-AWA} introduced an adaptive weight adjustment method that helps improve the performance of the original algorithm on complex Pareto fronts. 
The study proposed a two-stage strategy. First it introduces a novel weight vector initialization method based on a self-inverse transformation function, producing uniformly distributed solution mapping vectors. While this technique improves performance, it may struggle with complex Pareto fronts. To address this issue, the second stage adaptively adjusts weight vectors by removing subproblems from crowded regions and adding new ones to sparse areas, based on geometric analysis of the current Pareto-optimal solutions. Experiments on ZDT and DTLZ problems with up to 6 objectives showed that the proposed method performed better compared to standard MOEA/D, adaptive MOEA/D and NSGA-III, and obtained better uniformity at complex Pareto fronts.

Ying \textit{et. al} \cite{MOEAD-AU} introduced a method to improve MOEA/D's balance between convergence and diversity by incorporating the degree of the acute angles between reference directions and solutions as an additional selection criterion. For every new solution, the \textit{G} closest weight vectors in terms of acute angle are determined and the old solutions corresponding to these vectors are updated. This version was evaluated on 56 test problems including DTLZ and WFG with up to 10 objectives. Although the performance proved to be primarily governed by the parameter \textit{G}, the proposed algorithm achieved a better trade-off between convergence and diversity than the original MOEA/D and several of its variants. 

In another study aimed at improving population diversity of MOEA/D \cite{AES-MOEAD},  Wang \textit{et. al} proposed three modifications to the original algorithm. 
First, they used a combination of differential evolution (DE) and simulated binary crossover (SBX) interchangeably with a competitive polling mechanism. 
Tracking the population's statistics, every time the number of updating neighbors decreases---\textit{i.e} evolutions slows down---they replace DE with SBX. 
Second, they used an adaptive mutation probability that would reduce the probability of variation in the beginning of evolution when diversity is higher, and would increase it in the later stages. 
Finally, the standard boundary handling method forces out of bounds solutions to the nearest limit which leads to solutions crowding the boundaries. 
Instead, the authors proposed using a double-faced mirror reflection approach where the lower and upper limits are treated as mirrors, and out-of-bounds solutions are reflected back into the feasible range rather than fixed at the boundary. 
The effectiveness of these modifications was evaluated using ZDT and DTLZ benchmark suites with up to 10 objectives and the proposed algorithm showed better overall performance in terms of convergence and diversity compared to NSGA-II, SPEA-II \cite{SPEA-II}, PESA-II\cite{PESA-II}, and the original MOEA/D. 

\subsection{Lexicase Selection}
Even though Lexicase, as mentioned in Section II, was originally developed for genetic programming, previous research has shown it can perform well on multi- and many- objective optimization problems. La Cava \textit{et. al} \cite{elex} evaluated Lexicase on a set of regression benchmarks with 5 to 25 objectives and reported Mean Squared Error (MSE) on train and test sets. In an overall ranking of 9 algorithms across all datasets dynamic $\varepsilon$-lexicase had best performance among all other tested algorithms including lasso, age-fitness pareto optimization, and deterministic crowding. 

In another study, La Cava and Moore \cite{bill-poster} evaluated the performance of Lexicase on DTLZ problems with 5 to 100 objectives. They compared results to NSGA-II and Hype \cite{Hype} and concluded that Lexicase outperformed the two other algorithms in all experiments with more than 5 objectives, while NSGA-II is the best performer with 3 objectives. 

Finally, we previously \cite{gecco24} performed a theoretical analysis of lexicase and its variants on many-objective optimization problems. We proposed a theoretical test problem with up to 200 maximally conflicting objectives and identified a region of the parameter space in which Lexicase fails to identify Pareto-optimal solutions. However, outside of this region, we found that Lexicase is capable of successfully solving the problem and finding Pareto optimal solutions.

\section{Experimental setup}
The goal of this study is to compare the performance of state-of-the-art evolutionary algorithms on massive-objective optimization problems with up to 100 objectives. 
In particular, we are interested in how the interactions between objectives and parameters such as dimensionality and population size influence algorithm performance. 
To this end, we employed the DOSSIER diagnostic suite and evaluated algorithms using the Inverted Generational Distance (IGD) metric. 
IGD measures the average distance from each point on the reference Pareto front to its nearest solution in the set produced by the algorithm. This metric captures both convergence (proximity to the true front) and diversity (spread of solutions). 
Lower IGD values indicate better performance. Experiments were implemented using the Pymoo library \cite{pymoo} and executed for 5000 generations on each diagnostic problem. The reported results represent averages over 10 independent replicates.
During reproduction, all algorithms use the Simulated Binary Crossover (SBX) and Polynomial Mutation (PM) operators. The parameters for these genetic operators are chosen in accordance with the recommendations provided in \cite{NSGA3} and are summarized in Table II. The value of $\varepsilon$ in constant $\varepsilon$-lexicase selection is set to 0.5.

\begin{table}[h]
\centering
\caption{Parameter settings of recombination and mutation operators}
\begin{tabularx}{\linewidth}{X|c}
\hline
\textbf{Parameters} & \textbf{Value} \\
\hline
SBX probability, $p_c$ & 1 \\
Polynomial mutation probability, $p_m$ & $1/D$ \\
$\eta_c$ & 30 \\
$\eta_m$ & 20 \\
\hline
\end{tabularx}
\label{tab:params}
\end{table}

\section{Results}
\subsection{Analysis of Lexicase and Its Variants on Diagnostic Test Problems}
\begin{figure*}
\centering
\includegraphics[width=\textwidth]{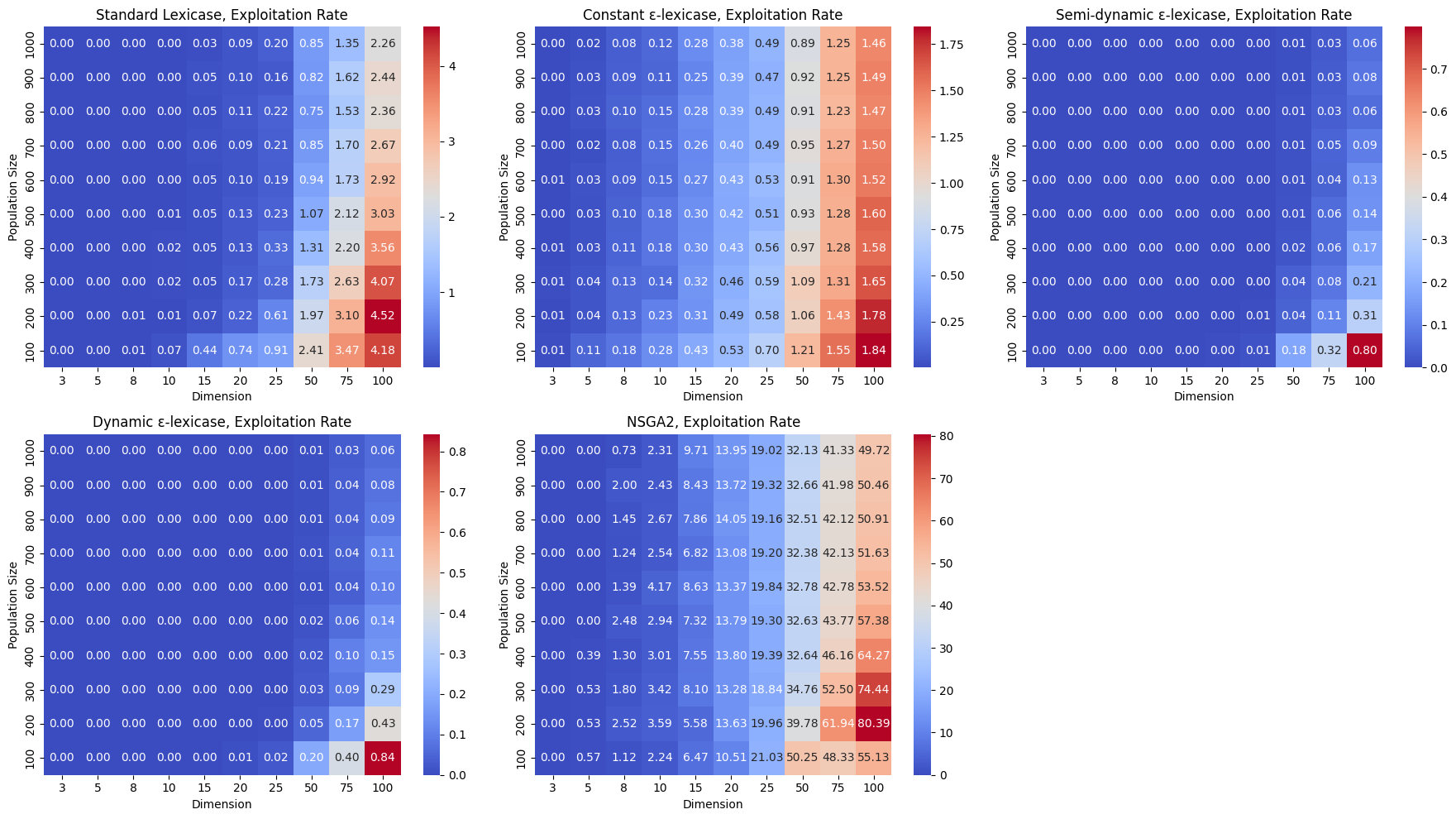}
\caption{Average IGD values in Exploitation Rate diagnostic for lexicase, $\varepsilon$-lexicase and its variants, and NSGA-II.}
\label{fig:exploit_hmap}
\end{figure*}

In this section, we analyze the performance of lexicase selection and its variants on DOSSIER diagnostic suite for a range of dimensions and population sizes. 
In our previous research we showed that lexicase selection is a good candidate for many-objective optimization \cite{gecco24}. Our goal here is to determine how lexicase selection compares with other algorithms designed specifically for multi- and many objective optimization. 
We test problems with varying numbers of objectives (ranging from 3 to 100) and population sizes (ranging from from 100 to 1000). 
We then compare the performance of lexicase selection and its variants with NSGA-II. 
We selected NSGA-II as a comparator for these experiments because it supports using arbitrary population sizes, making direct comparisons possible across the full parameter space.
In contrast, in reference-point-based algorithms, population size is dependent on the number of dimensions. 
Figure \ref{fig:exploit_hmap} shows results on the Exploitation Rate diagnostic across 5 algorithms: standard lexicase, constant $\varepsilon$-lexicase, semi-dynamic $\varepsilon$-lexicase and dynamic $\varepsilon$-lexicase. 
Results are reported in the form of heatmaps, where the x-axis shows the number of objectives and the y-axis shows population size. 
Colors correspond to average IGD values for each experiment. 

Based on the results presented in Figure 1, all variants of lexicase selection significantly outperform NSGA-II on the Exploitation Rate diagnostic, especially as the number of objectives increases. 
At the highest tested dimensionality (100 objectives), NSGA-II yields IGD values as high as 80 which is substantially higher (i.e. worse) than the IGD values achieved by any lexicase variant (IGD values are less than 5). 
Among the variants of Lexicase, semi-dynamic and dynamic $\varepsilon$-lexicase demonstrate comparable performance, both outperforming standard Lexicase and constant $\varepsilon$-lexicase. 
This observation suggests that when objectives are orthogonal, lexicase selection is a better choice than NSGA-II for problem solving. This result is not surprising since lexicase was designed for genetic programming purposes, where the test cases are unlikely to interact with each other, whereas NSGA-II was designed for problems with trade-offs.  

\begin{figure*}
\centering
\includegraphics[width=\textwidth]{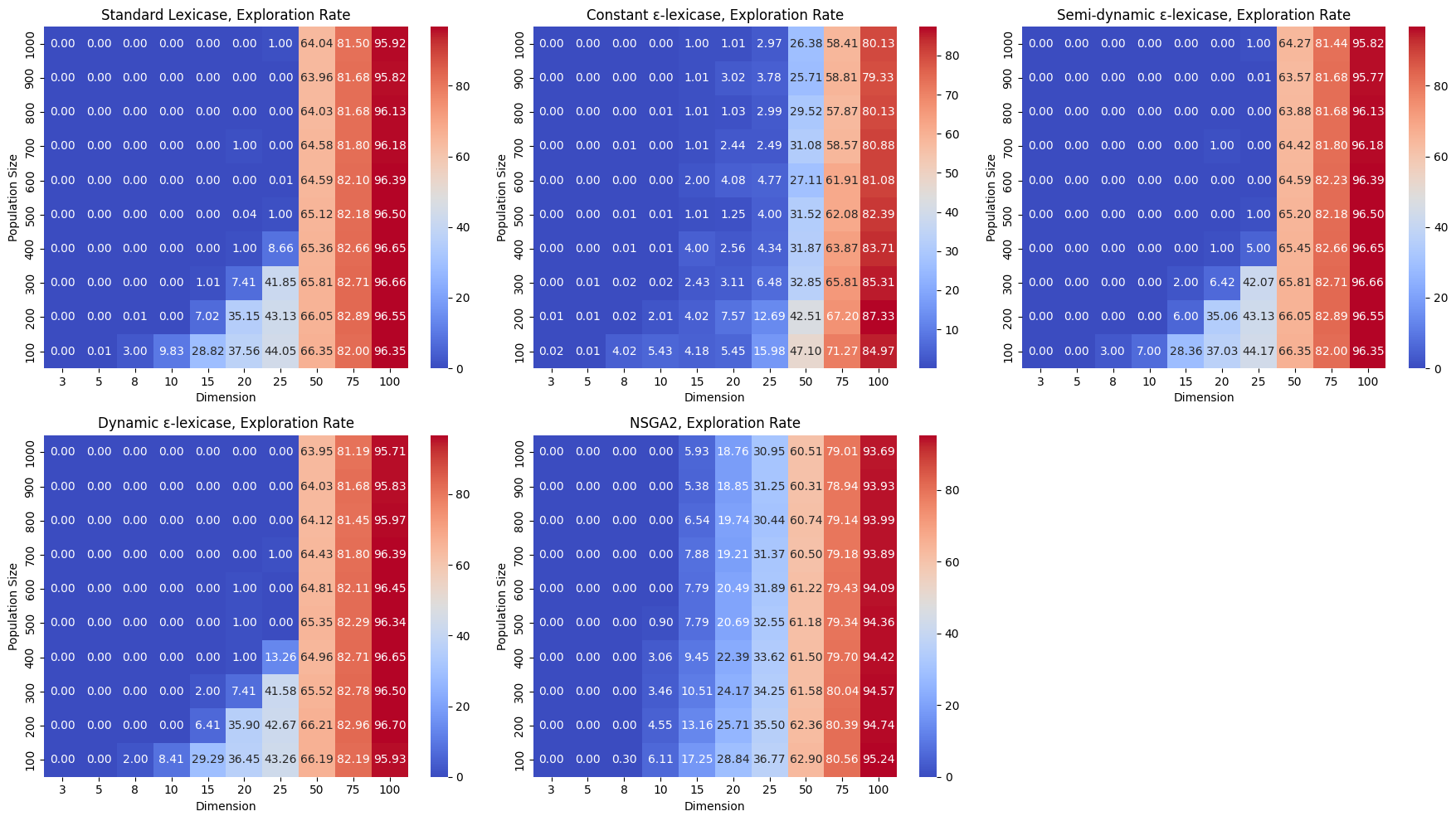}
\caption{Average IGD values in Multi-path Exploration Rate diagnostic for lexicase, $\varepsilon$-lexicase and its variants, and NSGA-II}
\label{fig:explore_hmap}
\end{figure*}

Figure \ref{fig:explore_hmap} shows results for the Multi-path Exploration Rate diagnostic. When the number of objectives is below 8, all tested algorithms successfully identify Pareto optimal solutions in most runs, resulting in IGD values that are either zero or near zero. As the number of objectives increases beyond 50, all algorithms begin to struggle and IGD values become high, with little variation across algorithms. In the intermediate range, between 10 and 25 objectives, lexicase and its variants outperform NSGA-II in most cases, specifically when population size is large. Among the variants of lexicase selection, standard, semi-dynamic and dynamic lexicase selection show almost similar performance in different regions of the parameter space. Constant $\varepsilon$-lexicase appears to outperform other variants in intermediate objective values with smaller population sizes. 

\begin{figure*}
\centering
\includegraphics[width=\textwidth]{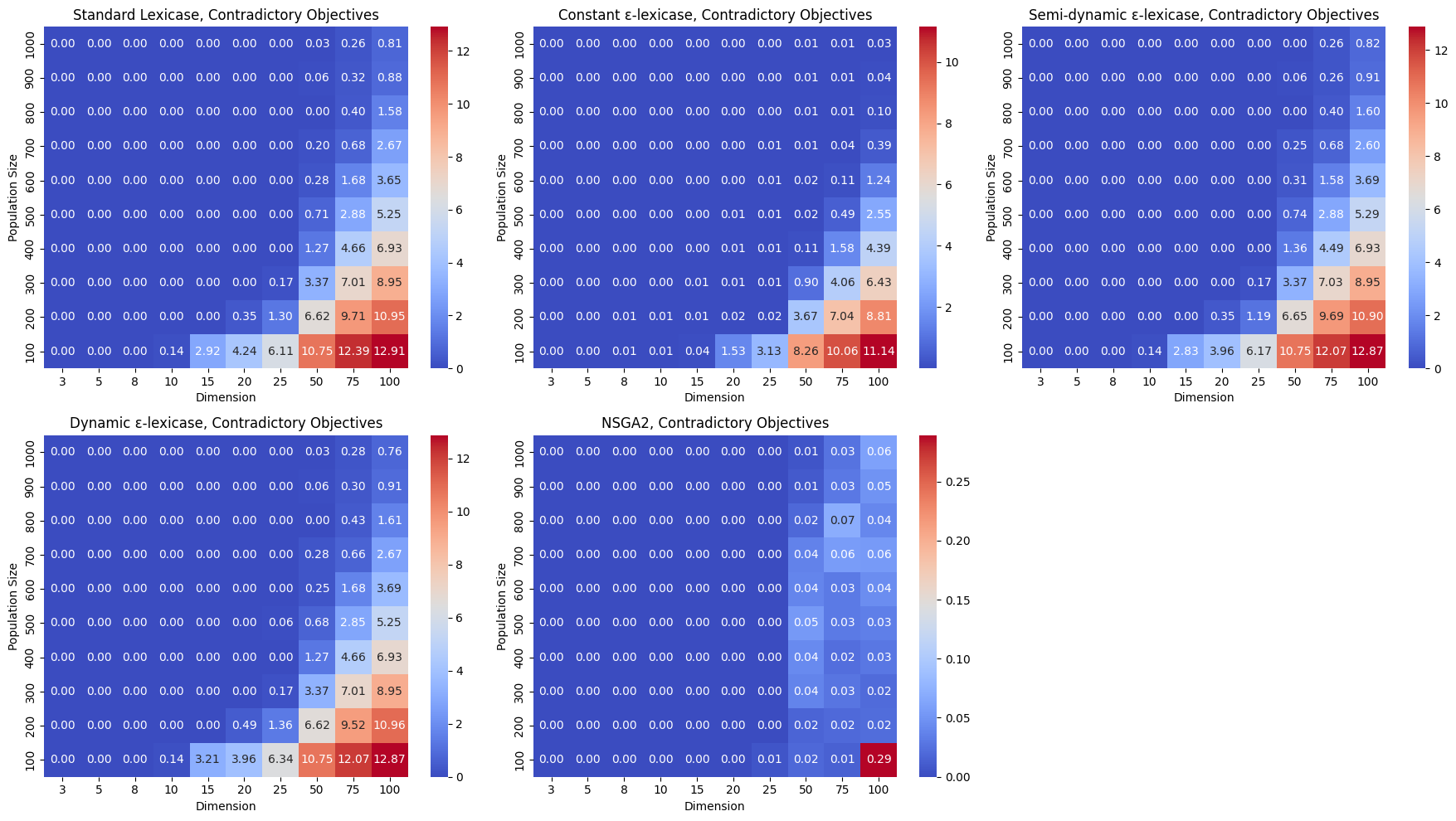}
\caption{Average IGD values in Contradictory Objectives diagnostic for lexicase, $\varepsilon$-lexicase and its variants, and NSGA-II}
\label{fig:weakDiversity_hmap}
\end{figure*}

Figure \ref{fig:weakDiversity_hmap} shows heatmaps for the Contradictory Objectives diagnostic, where objectives are conflicting with each other. In low-dimensional settings, all algorithms are able to approximate the reference Pareto front well, achieving low IGD values. However, as the number of objectives exceeds 15 and the population size remains below 500, lexicase shows a substantial decline in performance compared to NSGA-II. All variants of lexicase maintain relatively similar performance across the heatmap, with only minor differences observed among them. This result is consistent with the results of previous studies \cite{gecco24}.

\begin{figure*}
\centering
\includegraphics[width=\textwidth]{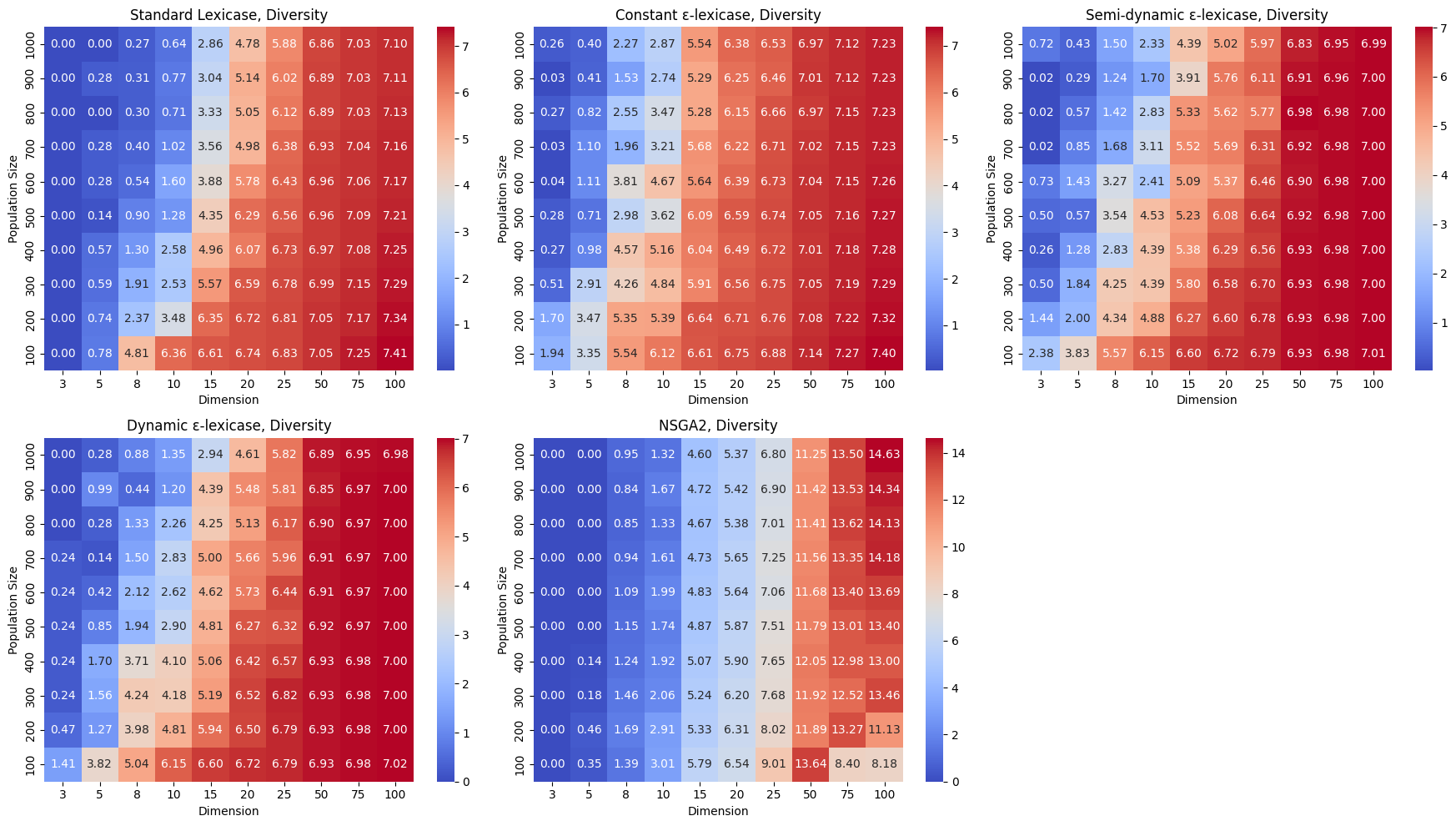}
\caption{Average IGD values in Diversity diagnostic for lexicase, $\varepsilon$-lexicase and its variants, and NSGA-II}
\label{fig:diversity_hmap}
\end{figure*}

Figure \ref{fig:diversity_hmap} shows a comparison of IGD values for the Diversity diagnostic, where the amount of conflicting and synergistic interactions are almost the same. For problems with fewer than 50 objectives, NSGA-II tends to achieve lower IGD values than lexicase and its variants, though the performance gap remains narrow. As the number of objectives exceeds 50, however, lexicase begins to significantly outperform NSGA-II, with IGD values becoming almost two times better. This difference is particularly noticeable when the population size approaches 1000, indicating that lexicase benefits more from larger population sizes in high-dimensional settings.

\begin{figure*}
\centering
\includegraphics[width=\textwidth]{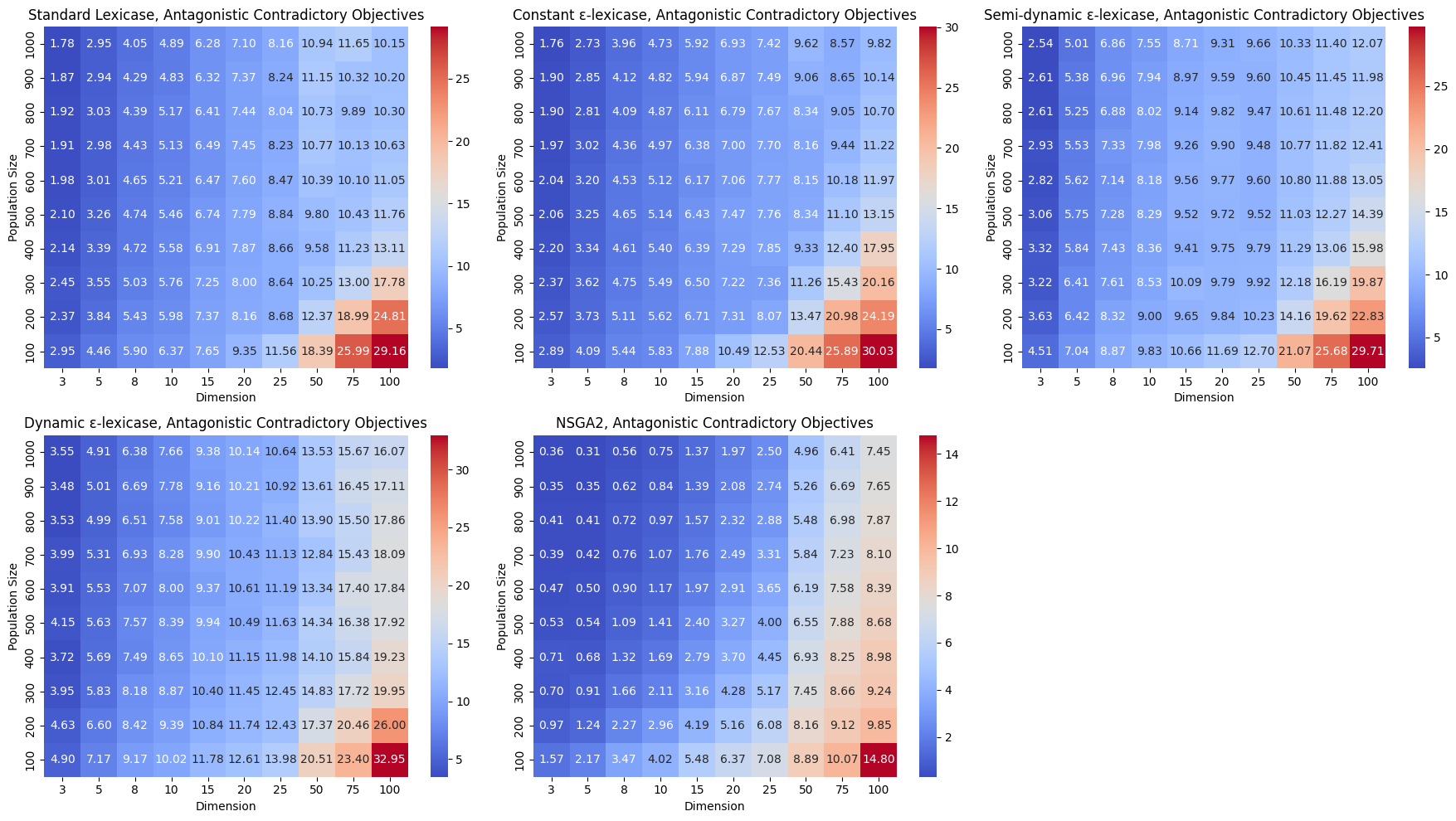}
\caption{Average IGD values in Antagonistic Contradictory Objectives diagnostic for lexicase, $\varepsilon$-lexicase and its variants, and NSGA-II}
\label{fig:antagonistic_hmap}
\end{figure*}

Finally, Figure \ref{fig:antagonistic_hmap} shows results for the Antagonistic Contradictory Objectives diagnostic, in which all objectives have maximal conflict with each other. Not surprisingly, results follow the same trend observed in Contradictory Objectives, with NSGA-II outperforming lexicase in the lower right region of the heatmaps where population size is low and dimensionality is high. This result is consistent with theoretical results in previous research \cite{gecco24}. One key distinction, however, is that unlike with Contradictory Objectives, none of the algorithms were able to achieve perfect IGD values. This pattern makes intuitive sense, as improving any objective in Antagonistic Contradictory Objectives degrades all other objectives, making the problem significantly more difficult to solve.

\subsection{Comparison of State-of-the-art Algorithms}

In this section, we investigate the performance of several state-of-the-art evolutionary multi and many-objective algorithms across different diagnostic test problems and dimensionalities. For NSGA-III and MOEA/D, the population size is determined by the number of reference directions (weight vectors), which is inherently dependent on dimensionality. As discussed in Section II, the population size in MOEA/D is equal to the number of reference points, while in NSGA-III it is set to the smallest multiple of four bigger than the number of reference points. For NSGA-II, we selected the population size that yielded the best performance for each dimensionality (selecting the smaller population in cases where multiple sizes performed equally). Similarly, for lexicase, we report results for the best-performing variant and population size for each dimension.

\begin{figure}[h]
\centering
\includegraphics[width=\linewidth]{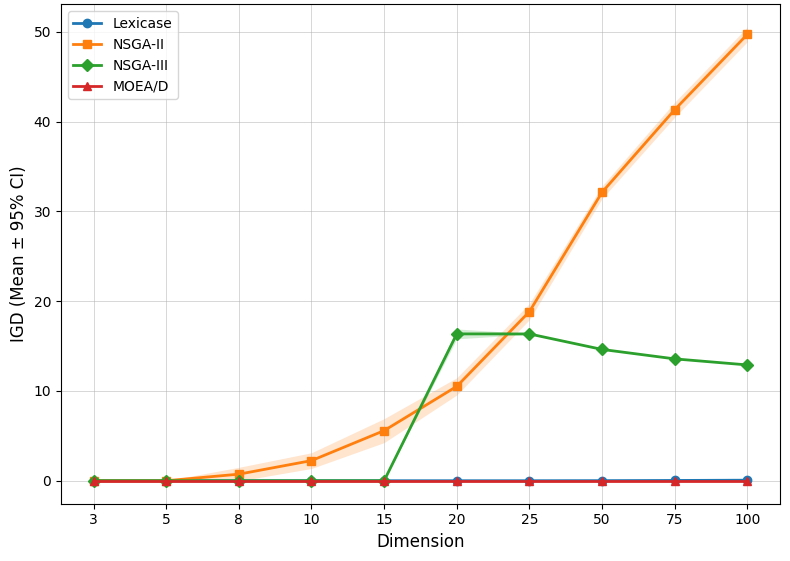}
\caption{Comparison of IGD values across algorithms in Exploitation Rate}
\label{fig:exploit_line}
\end{figure}

In Exploitation Rate, as shown in Figure \ref{fig:exploit_line} both MOEA/D and lexicase are able to find the Pareto optimal solution across all dimensionalities, leading to IGD values that are nearly zero. This behavior is expected from MOEA/D as the orthogonal nature of objectives allows the algorithm to decompose the problem into independent subproblems, each solvable in isolation. NSGA-III successfully locates the Pareto front for up to 15 objectives but is outperformed by lexicase and MOEA/D as the number of objectives increases. NSGA-II shows the weakest performance among the evaluated algorithms, with IGD values rising as high as 50 with 100 objectives. This figure shows that when objectives are orthogonal, lexicase selection or MOEA/D are obviously better choices for solving the problem. 

\begin{figure}[h]
\centering
\includegraphics[width=\linewidth]{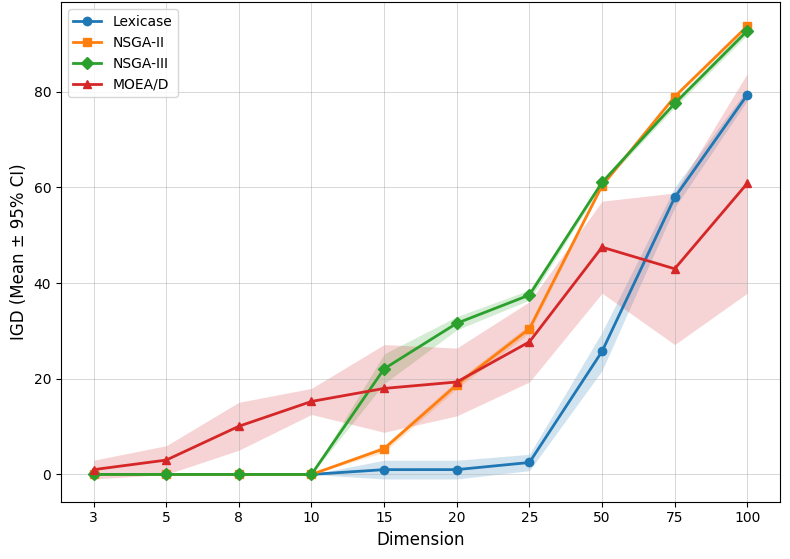}
\caption{Comparison of IGD values across algorithms in Exploration Rate}
\label{fig:explore_line}
\end{figure}

Figure~\ref{fig:explore_line} presents the results for the Multi-path Exploration diagnostic. Across all algorithms, IGD values increase with the number of objectives, indicating a general decline in performance as problem dimensionality grows. For problems with fewer than 10 objectives, all algorithms except MOEA/D are able to find the optimal Pareto front, achieving IGD values close to zero. As dimensionality increases, however, MOEA/D begins to outperform both NSGA-II and NSGA-III. 
Notably, lexicase performs significantly better than the other algorithms at a number of dimensionalities, while never performing worse than any algorithm. 

\begin{figure}[h]
\centering
\includegraphics[width=\linewidth]{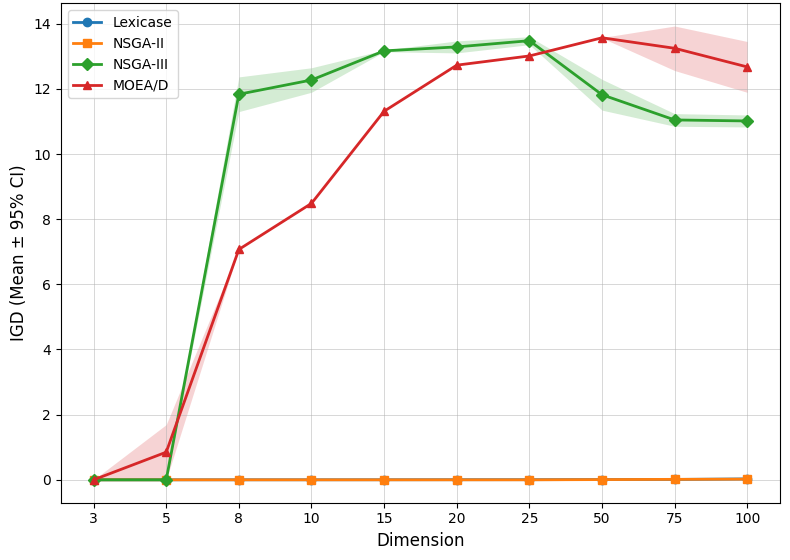}
\caption{Comparison of IGD values across algorithms in Contradictory Objectives}
\label{fig:weakDiversity_line}
\end{figure}

In Contradictory Objectives, lexicase and NSGA-II are the top 2 performers across all dimensionalities as shown in Figure \ref{fig:weakDiversity_line}. It is important to note, however, that lexicase requires a larger population size than NSGA-II in order to perform well. As an example, Figure \ref{fig:weakDiversity_hmap}) shows that at dimension 100, constant $\varepsilon$-lexicase with population size 900 performs as well as NSGA-II with population size 600. Surprisingly, NSGA-III and MOEA/D both show a significantly poorer performance across all but the two smallest dimensionalities, with MOEA/D performing slightly better at higher dimensions. 

\begin{figure}[h]
\centering
\includegraphics[width=\linewidth]{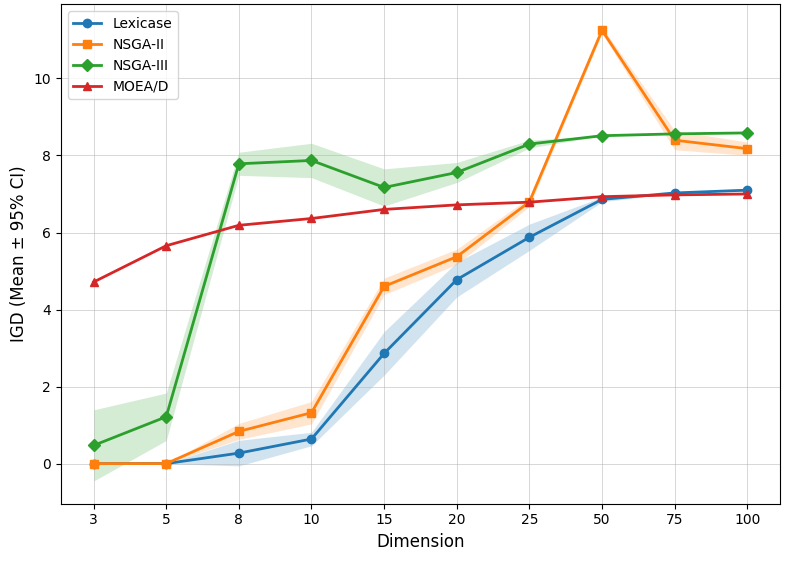}
\caption{Comparison of IGD values across algorithms in Diversity}
\label{fig:diversity_line}
\end{figure}

Figure \ref{fig:diversity_line} shows results for the Diversity diagnostic. While the performance patterns are more mixed in this test problem, lexicase seems to be the best performer across all dimensions. NSGA-II performs competitively when the number of objectives is below 25, but is overtaken by MOEA/D in higher dimensions. In this diagnostic, MOEA/D demonstrates little sensitivity to dimensionality, resulting in a nearly flat performance curve. Further analysis reveals that MOEA/D tends to converge to a single point on the reference Pareto front and remains trapped there, failing to explore other regions of the front.

\begin{figure}[h]
\centering
\includegraphics[width=\linewidth]{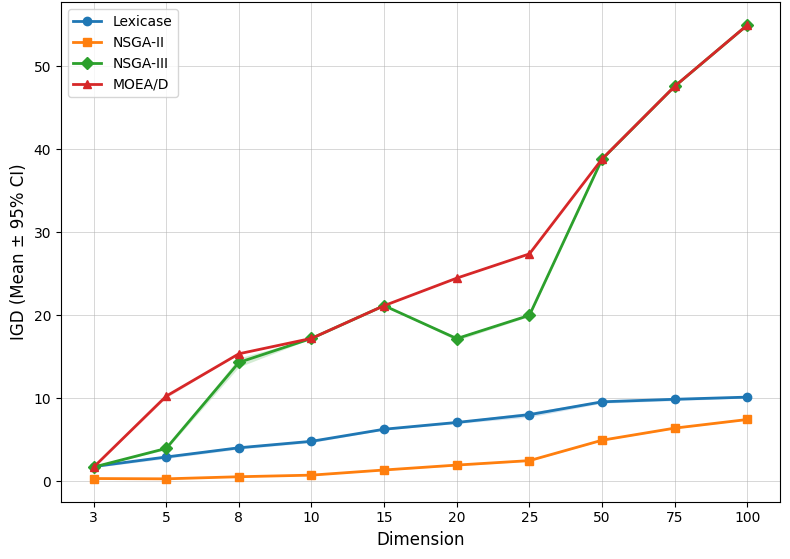}
\caption{Comparison of IGD values across algorithms in Antagonistic Contradictory Objectives}
\label{fig:antagonistic_line}
\end{figure}

Finally, in the Antagonistic Contradictory Objectives diagnostic, shown in Figure \ref{fig:antagonistic_line}, NSGA-II consistently outperforms all other algorithms across all dimensionalities. This result, along with its performance on the Contradictory Objectives diagnostic, highlights NSGA-II’s robustness in handling conflicting objectives, making it a strong choice in such problems regardless of the number of objectives. Lexicase follows a similar trend, with IGD values only slightly higher than those of NSGA-II. In contrast, both NSGA-III and MOEA/D show inferior performance, with IGD values increasing steadily as the number of objectives grows.

\section{Discussion and Conclusion}





In this study we investigated two fundamental questions that have remained largely unexplored despite decades of research in the field of many-objective optimization. First, how is the performance of different algorithms affected when the number of objectives becomes "massive", far beyond the conventional upper limit of approximately 15 objectives? Second, how do problem characteristics such as the interactions between the objectives influence the performance of these algorithms? To answer these questions, we employed a diagnostic benchmark suite that provides a range of controlled problems with known objective interactions while scaling the number of objectives to very high dimensions.

A key finding of this study is that different algorithms performed wildly differently on the different problems. We hypothesize that the relationships between objectives strongly influenced these results, although a more tightly controlled experiment would be required to confirm this hypothesis as it is possible that it was instead driven by other problem characteristics (\textit{e. g.} the topology of the fitness landscape).
Most prior work assumes that all objectives necessarily conflict with each other, but in practice objectives can also be -- and often are -- orthogonal or synergistic. Indeed, as the number of objectives gets large, it becomes highly improbable that they are completely conflicting.
The results revealed that in diagnostics where objectives have no conflict (such as in the Exploitation Rate diagnostic), lexicase selection and MOEA/D consistently outperform NSGA-II and NSGA-III.
In other diagnostics, MOEA/D’s effectiveness declines relative to lexicase, though it remains superior to NSGA-II and NSGA-III in certain dimensionalities and diagnostics.
Under the most extreme level of conflict, however, NSGA-II ultimately surpasses all other algorithms, even at the highest dimensions. These results highlight the fact that no single algorithm dominates across all types of objective relationships. Moreover, they provide guidance for which algorithm to select when faced with a given problem.

Another key finding is that lexicase selection, despite having been developed for genetic programming purposes, demonstrated competitive performance compared to state-of-the-art evolutionary algorithms designed for many objective optimization, such as NSGA-III and MOEA/D, across all diagnostics. Across all experiments, lexicase never did significantly worse than these algorithms, and sometimes performed dramatically better. It also outperformed NSGA-II on all diagnostics except for antagonistic contradictory objectives (where is pereformed only slightly worse). 
Lexicase selection also has the advantage over NSGA-III and MOEA/D of not requiring a predefined set of reference directions, a requirement that adds configuration complexity and increases the computational cost of adding objectives. These results demonstrate that lexicase selection is a promising algorithm for many-objective optimization. 

These findings are even more impressive in light of the fact that lexicase is a parent selection algorithm, unlike other methods in this paper which are full evolutionary algorithms with both parent selection and survival stages. For example, NSGA-II employs tournament selection for selecting parents, and NSGA-III uses random parent selection. Both algorithms then combine their selection method with a separate survival strategy, such as non-dominated sorting, to determine which individuals persist to the next generation. Lexicase, on the other hand, relies entirely on its parent selection procedure. 
This raises an interesting question for future work: could lexicase selection's performance be further improved if it were paired with a survival mechanism?

Another important thing to note is that, in our experiments, we used SBX crossover for recombination of selected parents. However, as suggested in \cite{NSGA3}, when the number of objectives becomes large, solutions become distant and the effectiveness of crossover becomes questionable. To examine this effect, we ran a set of experiments where we turned off the crossover operator. The results showed that crossover can make a difference in how algorithms perform, especially as the number of objectives increases. While NSGA-II and MOEA/D were minimally affected, Lexicase selection and NSGA-III showed more sensitivity to the absence of crossover. Additional details and figures are included in the supplemental material.

One limitation of this work is that all problems in the DOSSIER benchmark suite have the same number of variables as objectives. Future research should explore how independent scaling of the variables and objectives influences algorithmic performance. In addition, we only explored unconstrained many-objective optimization problems in this study, with simple bounds on variables. In future work, it would be worth looking at constrained problems in which trade-offs among objectives are explored within a predefined feasible region. Finally, none of the diagnostics used in this study include fitness landscapes with valleys.
Valleys are regions of low fitness that must be crossed to reach higher peaks, and can challenge evolutionary algorithms by increasing the risk of premature convergence to local optima. Future work could adopt the valley-crossing diagnostics introduced in \cite{diags} to examine how many-objective algorithms cope with such rugged landscapes.


Finally, we emphasize that the diagnostic fitness landscapes in this study represent Pareto front geometries that might differ from those typically assumed in classical multi- and many-objective optimization benchmarks. Conventional Pareto fronts are often characterized by smoothly varying trade-offs and a uniform distribution of solutions across the objective space, reflecting the common assumption that objectives conflict. In contrast, the Pareto fronts represented by our diagnostic landscapes are dominated by extreme solutions, with most non-dominated points concentrated near the axes or at corner regions of the search space.

This difference arises from the fact that many-objective problems. in practice, rarely consist of exclusively antagonistic objectives. As demonstrated in this work, objectives may have a mixture of interactions—including synergistic, orthogonal, and contradictory. We propose that such interactions give rise to Pareto fronts that deviate from the idealized, uniformly distributed fronts commonly used in benchmark studies on smaller numbers of objectives. A more rigorous study of the shapes of Pareto fronts in problems with very large numbers of objectives would help clarify this point.

Although the present work does not explicitly evaluate problems whose Pareto fronts resemble these more continuous trade-off surfaces, prior research has examined the behavior of lexicase selection under such conditions and found it to remain competitive \cite{elex}. These results suggest that the strengths of lexicase selection are not limited to the diagnostic landscapes in this study, but may generalize across a broader range of Pareto front geometries. A more systematic investigation of lexicase selection across a wider range of Pareto front geometries would be worthwhile, given the high degree of potential it has shown so far.

\section*{Acknowledgments}
We would like to acknowledge the members of the ECODE lab at Michigan State University for their feedback and support and Institute for Cyber-Enabled Research (ICER) for providing the computational power to do the experiments in this study.

\newpage

 





\end{document}